\documentclass{article} 
\usepackage{iclr2026_conference,times}

\usepackage{amsmath,amsfonts,bm}

\def\eqref#1{equation~\ref{#1}}

\def\1{\bm{1}}

\DeclareMathAlphabet{\mathsfit}{\encodingdefault}{\sfdefault}{m}{sl}
\SetMathAlphabet{\mathsfit}{bold}{\encodingdefault}{\sfdefault}{bx}{n}

\usepackage{hyperref}
\usepackage{url}
\usepackage{graphicx}
\usepackage{multicol}
\usepackage{booktabs}
\usepackage{amsmath}
\usepackage{multirow}
\usepackage{amssymb}
\usepackage{bbm}
\usepackage{subcaption}
\usepackage[capitalize]{cleveref}
\usepackage{wrapfig}
\usepackage{enumitem}
\usepackage{pgfplots}
\pgfplotsset{compat=1.18}
\usetikzlibrary{calc}

\title{Hierarchical Latent Action Model}

\author{
Hanjung Kim$^{1, 2}$, Lerrel Pinto$^{2}$, Seon Joo Kim$^{1}$\\
$^{1}$ Yonsei University, $^{2}$ New York University \\
}

\newcommand{\name}{HiLAM}

\iclrfinalcopy 
\begin{document}

\maketitle

\begin{abstract}
Latent Action Models (LAMs) enable learning from actionless data for applications ranging from robotic control to interactive world models.
However, existing LAMs typically focus on short-horizon frame transitions and capture low-level motion while overlooking longer-term temporal structure.
In contrast, actionless videos often contain temporally extended and high-level skills.
We present \name{}, a hierarchical latent action model that discovers latent skills by modeling long-term temporal information.
To capture these dependencies across long horizons, we utilize a pretrained LAM as a low-level extractor.
This architecture aggregates latent action sequences, which contain the underlying dynamic patterns of the video, into high-level latent skills.
Our experiments demonstrate that \name{} improves over the baseline and exhibits robust dynamic skill discovery.
\end{abstract}

\section{Introduction}
Recent progress in robot learning has increasingly relied on incorporating large-scale data for training.
However, obtaining action-labeled data is prohibitively expensive and makes it difficult to ensure dataset diversity.
To remedy this, Latent Action Models~\citep{lapo, genie, lapa, uniskill} have emerged as a prominent approach by extracting latent actions directly from observation-only data.
Generally, by utilizing inverse and forward dynamics models, LAMs infer the latent action between two frames.
These latent actions are then used for pretraining Vision-Language-Action models (VLAs) with actionless data~\citep{lapa, univla}, transferring actions across different embodiments~\citep{uniskill}, or enabling interaction within world models~\citep{genie, adaworld}.

Latent actions offer a promising way to leverage dynamic information from diverse video sources.
Despite their flexibility, existing latent action models are largely limited to short-term motion.
As a result, they can capture low-level dynamics from observation-only data but often miss higher-level structure, such as temporally extended skills.
This exposes a key gap where actionless videos contain not only primitive motions but also high-level skills that remain underutilized.

This raises a natural question: how can we extract such skills from unlabeled video?
Prior work typically either assumes a fixed set of skill vectors~\citep{buds, skilldiffuser} or encodes fixed-length sequences of low-level actions into skill representations~\citep{spirl}.
In contrast, real-world skills vary in duration, and large-scale data introduces an increasingly diverse set of behaviors.
Even for the same task, demonstrations can vary substantially in execution speed and, consequently, in skill duration.
When skills are forced into a fixed-length window, two trajectories that express the same underlying behavior may be mapped to very different skill representations.
Another line of work uses language to define skills~\citep{hirobot}.
However, it typically segments behavior from task descriptions, such as by splitting an instruction into sub-instructions, rather than from motion cues.
Therefore, language is a complementary signal for skill discovery and not a replacement for modeling dynamics.


To this end, we propose \name{}, a hierarchical latent action model that encodes latent skills from sequences of latent actions, regardless of their length or the need for pre-defined skill sets.
\ref{fig:fig1} demonstrates the overall architecture of \name{}.
To enable a dynamic hierarchical design, we adopt the H-Net~\citep{hnet} architecture, which introduces a novel dynamic chunking mechanism that automatically segments boundaries.
Following the H-Net framework, we formulate \name{} using a next-token prediction approach during pretraining, utilizing latent actions extracted from an inverse dynamics model (IDM).
Additionally, predicted latent actions are used to reconstruct future frames, consistent with prior works~\citep{lapa, uniskill}, to maintain their dynamic motion properties.
Due to H-Net's dynamic chunking mechanism, latent actions are naturally grouped into similar representations of varying lengths without the need for action labels.
These representations are then processed through an encoder to serve as latent skills.
Once these skills are obtained, we train a skill policy to predict the latent skill based on the current observation.
Simultaneously, we train a skill-conditioned policy to predict low-level actions based on the observation and the predicted latent skills.

Our experiments show that \name{} is able to detect and encode skill representations while remaining free from constraints on length or pre-defined skill sets.
Furthermore, the predicted next latent actions maintain interpretability, which is demonstrated by predicting future frames corresponding to the given latent actions.
In terms of training computation, since \name{} reuses pretrained LAMs for extracting latent actions, it is capable of encoding long-horizon trajectories efficiently.

\section{Related Work}
\subsection{Latent action learning}
Latent action learning is a prominent approach for inferring actions from observation-only data.
By analyzing frame transitions, Latent Action Models (LAMs) extract the latent action between frames using an Inverse Dynamics Model (IDM).
LAPO~\citep{lapo} focuses on inferring discrete latent actions from gaming environments, while Genie~\citep{genie} proposes an interactive world model for games through a discrete latent action space.
Since prior works were largely limited to simulated environments, LAPA~\citep{lapa} introduces latent action learning to robotics, leveraging diverse actionless data by utilizing latent actions as pseudo-labels for training VLAs.

Standard LAMs often utilize a Forward Dynamics Model (FDM) to predict future images, where the reconstruction objective encodes dynamic information into the latent action.
However, this process can accidentally incorporate task-irrelevant information into latent representation.
To address this, UniVLA~\citep{univla} proposes a task-centric learning approach, while UniSkill~\citep{uniskill} adopts an image-editing pipeline and LAOM~\citep{laom} utilizes supervised learning to reduce such dependencies.
Furthermore, to represent more diverse action spaces, UniSkill~\citep{uniskill} and CLAM~\citep{clam} employ continuous latent actions rather than discrete ones. 
Latent actions also serve as a substitute for explicit action labels in world models, which is necessary for interaction.
Recent works such as AdaWorld~\citep{adaworld} and Latent Action World Model~\citep{garrido2026learning} utilize latent action learning specifically for world model training.

However, these LAMs are generally limited to short-period motion patterns and lack the capacity to represent high-level skills.
To address this, \name{} introduces a method for extracting latent skills from actionless data.

\subsection{Hierarchical Robot Learning}
Hierarchical robot learning uses skill representations as an intermediate abstraction for action prediction, enabling policies to better handle long-horizon and complex tasks.
Unlike low-level actions, skills are difficult to annotate manually within demonstrations.
Consequently, prior work often adopts unsupervised training paradigms to learn these abstractions from action sequences.

SPiRL~\citep{spirl} and SkiLD~\citep{skild} employ autoencoder architectures to learn skill representations from fixed-length action sequences, using these learned priors to accelerate reinforcement learning in downstream tasks.
SAILOR~\citep{sailor} also uses an autoencoder framework for skill extraction, further incorporating a temporal-distance objective to ensure the learned representations are temporally aware.
Rather than using latent embeddings, BUDS~\citep{buds} discovers skills by clustering unsegmented demonstrations into temporal segments based on a set of predefined skill primitives.
SkillDiffuser~\citep{skilldiffuser} adopts VQ-VAE~\citep{vqvae} to encode high-level instructions into a discrete set of learnable skills and uses them for future-frame generation.
Meanwhile, Hi Robot~\citep{hirobot} uses a high-level VLM to map user prompts into low-level language commands.

Prior work typically assumes a fixed horizon for encoding skill representations, a fixed number of skill primitives, or a different modality.
In contrast, \name{} introduces a dynamic chunking mechanism that abstracts sequences of low-level motions in a data-driven and length-adaptive way.

\section{Method}
In this paper, we propose \name{}, a hierarchical latent action model.
The framework consists of two phases.
In the first phase, \name{} is trained to predict the next latent action utilizing a dynamic chunking mechanism.
Due to its architectural properties, \name{} automatically detects skill boundaries within untrimmed input latent action sequences.
In the second phase, we utilize these latent actions and the encoded latent skills to train high-level and low-level policies.
Since each latent skill contains shared information within a chunked segment, the high-level policy is trained to predict latent skills, while the low-level policy is trained to predict latent actions.
Finally, we fine-tune the low-level policy with ground-truth actions to map the latent action space to the true action space.

\subsection{Preliminaries}
\subsubsection{Problem Formulation}
\begin{figure}[t]
    \centering
    \includegraphics[width=\textwidth]{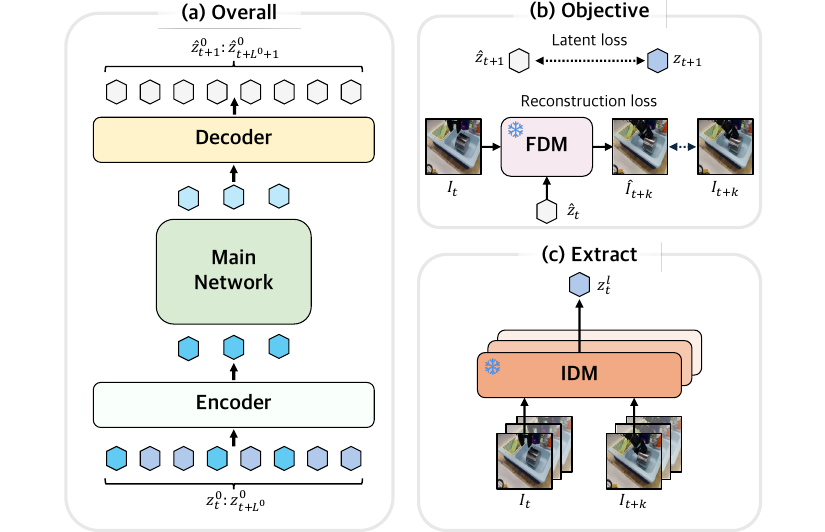}
    \caption{Overview of \name{}. (a)~Overall latent skill learning pipeline. (b)~Training objectives used for latent skill learning. (c)~Extracting latent actions using a pretrained inverse dynamics model (IDM).}
    \label{fig:fig1}
\end{figure}

We formalize the objective as encoding high-level latent skills, $z^h$, from observation-only videos $\mathcal{V}$ to facilitate hierarchical policy learning.
Our approach decomposes long-horizon trajectories into a hierarchy of latent representations, mapping visual observations to executable actions.
Given an observation-only video $\mathcal{V}$ of length $T$, we first extract a sequence of low-level latent actions $\{z^l_1, ..., z^l_{T-k}\}$.
Following prior work~\citep{lapa, uniskill}, we use an inverse dynamics model to infer the motion between two frames, $I_t$ and $I_{t+k}$, separated by a fixed temporal interval $k$ (\Cref{fig:fig1}c).
To capture temporal abstractions, this sequence of low-level latent actions is further compressed into a sequence of high-level latent skills $\{z^h_1, ..., z^h_S\}$, where $S < T-k$.
We allow both $T$ and $S$ to be variable, accommodating trajectories and skills of varying temporal durations.
For downstream control, we employ a hierarchical policy framework.
At each decision step, a high-level policy observes the current state $o_t$ and a task instruction $l$ to predict a target latent skill, $z^h_t\sim\pi^h(z^h_t|o_t, l)$.
Subsequently, conditioned on this high-level skill and the current observation, a low-level policy generates the primitive action $a_t$ for robot execution via $a_t\sim\pi^l(a_t|o_t, z^h_t)$.

\subsubsection{Dynamic Chunking Mechanism}
To abstract long sequences of latent actions into temporally extended skills, \name{} adopts the H-Net~\citep{hnet} architecture, which learns a data-driven segmentation of the input via dynamic chunking.
At a given stage $s$, let the input sequence be $\mathbf{z}^s = (z^s_1,\dots,z^s_{L^s})$.
An encoder $\mathcal{E}^s$ maps each token to a feature vector $h^s_t \in \mathbb{R}^d$.
H-Net then predicts boundary indicators $b^s_t \in \{0,1\}$ that decide whether token $t$ starts a new chunk.
We interpret $b^s_t=1$ as a \emph{segment-start} boundary (i.e., token $t$ is the first token of a new chunk).
Following~\citet{hnet}, we compute normalized query/key features $\hat{\mathbf{q}}^s_t$ and $\hat{\mathbf{k}}^s_t$ from $h^s_t$ and define
\begin{equation}
\label{eq:boundary}
    p^s_t =
    \begin{cases}
        1, & t = 1,\\
        \frac{1}{2}\bigl(1 - (\hat{\mathbf{q}}^s_{t-1})^\top \hat{\mathbf{k}}^s_t\bigr), & t > 1,
    \end{cases}
    \qquad
    b^s_t = \mathbbm{1}_{\{p^s_t \ge 0.5\}}.
\end{equation}
Intuitively, $p^s_t$ is large when consecutive tokens are dissimilar, encouraging a boundary at $t$.
Given the boundary pattern, we perform chunking (downsampling) by selecting only the boundary features.
Let $\mathcal{I}^s = \{t \mid b^s_t = 1\}$ denote the selected boundary indices (ordered increasingly), and let $L^{s+1} = |\mathcal{I}^s|$.
The chunked sequence is obtained as
\begin{equation}
    z^{s+1}_i = h^s_{t_i}, \qquad \text{where } t_i \in \mathcal{I}^s.
\end{equation}
That is, the stage-$(s+1)$ tokens are selected encoder features at boundary indices and serve as chunk-level summaries.
A main network $\mathcal{M}^s$ then processes the shorter sequence $\mathbf{z}^{s+1} = (z^{s+1}_1,\dots,z^{s+1}_{L^{s+1}})$, and a decoder $\mathcal{D}^s$ expands the processed sequence back to length $L^s$ conditioned on the same boundary pattern.
Overall, the encode--chunk--main--dechunk stage is summarized by
\begin{equation}
\begin{aligned}
    \mathbf{h}^s &= \mathcal{E}^s(\mathbf{z}^s), \qquad & \mathbf{z}^{s+1} &= \mathrm{Chunk}(\mathbf{h}^s; \mathbf{b}^s), \\
    \hat{\mathbf{z}}^{s+1} &= \mathcal{M}^s(\mathbf{z}^{s+1}), \qquad & \hat{\mathbf{z}}^{s} &= \mathrm{DeChunk}(\mathcal{D}^s(\hat{\mathbf{z}}^{s+1}); \mathbf{b}^s).
\end{aligned}
\label{eq:dc}
\end{equation}
Stacking multiple stages yields a hierarchical representation in which higher levels operate on progressively shorter, chunk-level sequences.

\subsection{Latent Skill Learning}
\label{sec:latent_skill}
We learn a hierarchy of latent skill representations from observation-only videos using a hierarchical sequence model.
In sequence modeling, inputs typically consist of language tokens, DNA bases, or action trajectories.
However, because ground-truth actions are unavailable in observation-only data, we use a pretrained Inverse Dynamics Model (IDM)~\citep{lapa, uniskill} to extract latent actions.
We then apply the dynamic chunking mechanism~\citep{hnet} to segment the latent action sequence into meaningful temporal chunks, encoding each chunk as a latent skill (\Cref{fig:fig1}).

Let $\mathbf{z}^l$ denote the resulting latent action sequence, and we use it as the initial input to \name{}, i.e., $\mathbf{z}^0 = \mathbf{z}^l$.
After each encoder $\mathcal{E}^s$, the model predicts boundary indicators $\mathbf{b}^s$ and selects representative tokens to form the chunked sequence $\mathbf{z}^{s+1}$.
As described in \Cref{eq:boundary}, boundaries are determined from feature dissimilarities, encouraging segmentation at points of large temporal change.
The selected tokens summarize each segment, effectively compressing a variable-length sequence of latent actions into a shorter sequence of segment-level representations.
We treat these higher-level tokens $\mathbf{z}^s$ (for $s>0$) as latent skill representations, denoted $\mathbf{z}^h$.

Finally, after hierarchical processing through the encoder--main--decoder stack, \name{} predicts the next latent actions at the lowest level.
Given an input sequence $\mathbf{z}^0 = (z^0_1,\dots,z^0_{L^0})$, the model outputs $\hat{\mathbf{z}}^0 = (\hat{z}^0_2,\dots,\hat{z}^0_{L^0+1})$ via next-token prediction.

\paragraph{Training objective.}
We optimize a weighted combination of (i) next-latent prediction, (ii) a visual supervision term that can be instantiated in different ways, and (iii) the H-Net chunking regularizer:
\begin{equation}
    \mathcal{L} = \mathcal{L}_{\mathrm{latent}} + \lambda_{\mathrm{rec}}\,\mathcal{L}_{\mathrm{rec}} + \lambda_{\mathrm{ratio}}\,\mathcal{L}_{\mathrm{ratio}}.
\end{equation}
In this formulation, $\mathcal{L}_{\mathrm{latent}}$ is an element-wise $\ell_1$ loss~\citep{vjepa, vjepa2} between predicted and target latent actions, modeling the next-token prediction task.
The term $\mathcal{L}_{\mathrm{rec}}$ provides additional supervision to ensure that the predicted latents maintain their action-like properties.
To achieve this, we employ a pretrained Forward Dynamics Model (FDM) to predict future frames conditioned on the predicted latent actions.
For instance, if $\hat{z}^0_{t+1}$ is predicted from $z^0_{t}$, the FDM is expected to predict the future frame $\hat{I}_{t+k+1}$ using the current frame $I_{t+1}$ and the predicted latent action $\hat{z}^0_{t+1}$.
To preserve the dynamic motion properties of the latent action, we define $\mathcal{L}_{\mathrm{rec}}$ as the reconstruction error between the predicted frame $\hat{I}_{t+k+1}$ and the ground-truth frame $I_{t+k+1}$.
Finally, $\mathcal{L}_{\mathrm{ratio}}$ denotes the H-Net ratio regularizer~\citep{hnet} that prevents degenerate boundary patterns and controls the average chunk length.

\paragraph{Latent Skill Extraction.}
After training \name{}, we extract stage-wise latent skills and align them to the original video length $T$ for downstream hierarchical policy learning.
Given an observation-only video $\mathcal{V}=\{I_1,\dots,I_T\}$, \name{} produces at each stage $s$ a downsampled sequence of latent representations $\mathbf{z}^s=\{z^s_1,\dots,z^s_{L^s}\}$ with $L^s \le T$ (\Cref{eq:dc}).
Because $\mathbf{z}^s$ is defined at the stage resolution, we expand it back to the original temporal resolution using the unfolded boundary.

\begin{figure}[t]
    \centering
    \includegraphics[width=\textwidth]{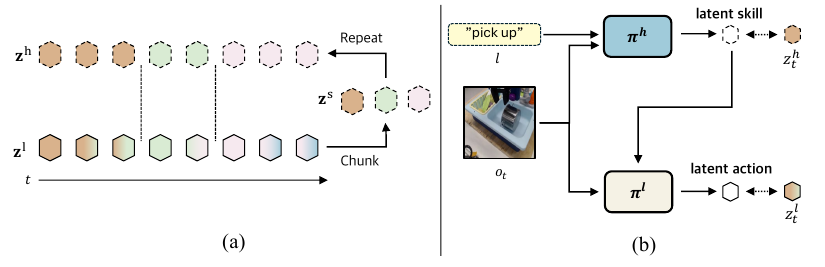}
    \caption{Latent skill extraction and policy learning. (a)~Latent actions $\mathbf{z}^l$ are hierarchically encoded into stage-wise representations $\mathbf{z}^s$ and then expanded back to a per-timestep latent skill sequence $\mathbf{z}^h$. (b)~Overall pipeline of the hierarchical skill policy.}
    \label{fig:fig2}
\end{figure}

Let $\bar{\mathbf{b}}^s\in\{0,1\}^T$ denote the \emph{unfolded boundary} of stage $s$, i.e., the boundary pattern expressed at the original temporal resolution.
We assign each timestep $t$ to its segment ID via a cumulative sum and expand the stage-$s$ sequence by indexing:
\begin{equation}
\label{eq:unfold_map}
k^s_t \;=\; \sum_{\tau=1}^{t} \bar{b}^s_\tau,
\qquad
\bar{z}^s_t \;=\; z^s_{k^s_t},
\qquad t=1,\dots,T.
\end{equation}
Here, $k^s_t$ is constant within a segment and increases by $1$ whenever $\bar{b}^s_t=1$, so each timestep inherits the latent representation of its corresponding segment.
As shown in \Cref{fig:fig2}, the resulting sequence $\bar{\mathbf{z}}^s=\{\bar{z}^s_1,\dots,\bar{z}^s_T\}$ serves as the per-timestep latent skill sequence $\mathbf{z}^h$ used for downstream policy learning.

\subsection{Hierarchical Policy Learning}
To leverage the learned latent skills for control, we train a hierarchical policy consisting of a high-level policy $\pi^h$ and a low-level policy $\pi^l$ (\Cref{fig:fig2}b).
We consider two training phases: pretraining on large-scale, actionless videos and fine-tuning on a target domain with action labels.

\paragraph{Pretraining.}
During pretraining, we supervise both policies using the latent skill/action sequences extracted by \name{}.
The high-level policy predicts a latent skill from the current observation and task description,
$\hat{z}^h_t \sim \pi^h(z^h_t\mid o_t, l)$.
Conditioned on the observation and the predicted skill, the low-level policy predicts the corresponding latent action,
$\hat{z}^l_t \sim \pi^l(z^l_t\mid o_t, \hat{z}^h_t)$.
We train $\pi^h$ and $\pi^l$ to match the extracted targets $z^h_t$ and $z^l_t$, respectively.
Because these targets are obtained from observation-only videos, pretraining can use diverse video sources (e.g., robot or human videos).

\paragraph{Fine-tuning.}
After pretraining, we freeze the high-level policy $\pi^h$ and fine-tune the low-level policy on target-domain data with ground-truth actions.
Given the predicted skill $\hat{z}^h_t$, the low-level policy outputs a real action for execution,
$\hat{a}_t \sim \pi^l(a_t\mid o_t, \hat{z}^h_t)$.

\section{Experiments}
\subsection{Experiment Setup}
\paragraph{Datasets}
We train \name{} on observation-only video datasets spanning both human and robot behaviors.
For human videos, we use Something-Something V2~\citep{sthsth}, which contains short clips of humans performing diverse object-centric actions.
For real-world robot videos, we use Droid~\citep{droid} and BridgeV2~\citep{bridge}, which are large-scale datasets collected with Franka and WidowX-250 robot arms, respectively.

\paragraph{Implementation details}
Following H-Net~\citep{hnet}, \name{} uses a two-stage H-Net.
To extract latent actions from actionless videos, we use UniSkill~\citep{uniskill}'s inverse dynamics model (IDM) as the latent-action tokenizer, and its forward dynamics model (FDM) for frame prediction conditioned on the predicted latent actions.
For latent skills $\mathbf{z}^h$, we use the stage-$s=2$ representations, i.e., $\mathbf{z}^h \equiv \mathbf{z}^2$.
For hierarchical policy learning, both the high-level and low-level policies are based on the BAKU~\citep{baku} architecture, and we use a T5 encoder~\citep{2020t5} as the language encoder.
For pretraining, we use either human videos (Something-Something V2) or robot videos (BridgeV2).
In both cases, we treat the data as observation-only: we discard any available action annotations, extract latent actions/skills solely from each video, and use them as pseudo-labels to train both the high-level and low-level policies.
For fine-tuning, we freeze the high-level policy and train only the low-level policy using expert demonstrations.
Unless otherwise stated, both pretraining and fine-tuning are run for 100k gradient steps.

\paragraph{Benchmark}
We evaluate downstream control on the LIBERO benchmark~\citep{libero}.
We report results on four suites: LIBERO-Spatial, LIBERO-Object, LIBERO-Goal, and LIBERO-Long.
LIBERO-Spatial emphasizes spatial reasoning, while LIBERO-Object tests generalization by varying the manipulated objects.
LIBERO-Goal uses consistent objects and backgrounds, requiring the policy to follow the language instruction to succeed.
LIBERO-Long is the most challenging suite, consisting of long-horizon tasks with multiple sub-goals.
Each suite contains 10 tasks, and each task provides 50 demonstration trajectories.
We fine-tune on the provided expert demonstrations and report success rates using the official evaluation rollouts.

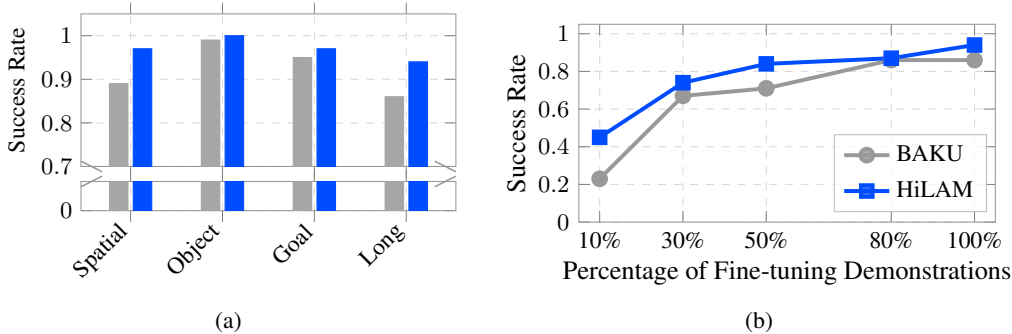
\begin{figure}[b]
    \centering
    \begin{subfigure}[t]{0.47\linewidth}
        \centering
        \small
        \begin{tikzpicture}
            \begin{axis}[
                name=axlow,
                ybar,
                bar width=7pt,
                width=\linewidth,
                height=0.30\linewidth,
                ymin=0, ymax=0.6,
                symbolic x coords={Spatial,Object,Goal,Long},
                xtick=data,
                x tick label style={rotate=45, anchor=east},
                ytick={0, 0.7},
                grid=major,
                grid style={dashed, gray!30},
                axis line style={gray!80},
                tick label style={font=\small},
                enlarge x limits=0.18,
            ]
                \addplot[fill=gray!70, draw=gray!70]
                coordinates {(Spatial,0.89) (Object,0.99) (Goal,0.95) (Long,0.86)};
                \addplot[fill=blue!70!cyan, draw=blue!70!cyan]
                coordinates {(Spatial,0.97) (Object,1.00) (Goal,0.97) (Long,0.94)};
            \end{axis}

            \begin{axis}[
                name=axhigh,
                at={(axlow.north west)},
                anchor=south west,
                yshift=2mm,
                ybar,
                bar width=7pt,
                width=\linewidth,
                height=0.55\linewidth,
                ymin=0.7, ymax=1.05,
                ylabel={Success Rate},
                symbolic x coords={Spatial,Object,Goal,Long},
                xtick=data,
                xticklabels=\empty,          
                ytick={0.7, 0.8, 0.9, 1.0},
                grid=major,
                grid style={dashed, gray!30},
                axis line style={gray!80},
                tick label style={font=\small},
                enlarge x limits=0.18,
            ]
                \addplot[fill=gray!70, draw=gray!70]
                coordinates {(Spatial,0.89) (Object,0.99) (Goal,0.95) (Long,0.86)};
                \addplot[fill=blue!70!cyan, draw=blue!70!cyan]
                coordinates {(Spatial,0.97) (Object,1.00) (Goal,0.97) (Long,0.94)};
            \end{axis}

        \draw[gray!80, line width=0.6pt]
            ($(axlow.north west)+(0pt,-2pt)$) -- ($(axlow.north west)+(8pt,2pt)$);
        \draw[gray!80, line width=0.6pt]
            ($(axhigh.south west)+(0pt,2pt)$) -- ($(axhigh.south west)+(8pt,-2pt)$);

        \draw[gray!80, line width=0.6pt]
            ($(axlow.north east)+(-8pt,-2pt)$) -- ($(axlow.north east)+(0pt,2pt)$);
        \draw[gray!80, line width=0.6pt]
            ($(axhigh.south east)+(-8pt,2pt)$) -- ($(axhigh.south east)+(0pt,-2pt)$);

        \end{tikzpicture}
        \caption{ }
        \label{fig:libero_suites}
    \end{subfigure}
    \hfill
    \begin{subfigure}[t]{0.52\linewidth}
        \centering
        \begin{tikzpicture}
            \begin{axis}[
                width=0.98\linewidth,
                height=0.58\linewidth,
                xlabel={Percentage of Fine-tuning Demonstrations},
                ylabel={Success Rate},
                ylabel style={yshift=-2pt},
                xlabel style={yshift=2pt},
                xmin=5, xmax=105,
                ymin=0, ymax=1.05,
                xtick={10,30,50,80,100},
                xticklabels={10\%, 30\%, 50\%, 80\%, 100\%},
                ytick={0, 0.2, 0.4, 0.6, 0.8, 1.0},
                grid=major,
                grid style={dashed, gray!30},
                axis line style={gray!80},
                tick label style={font=\small},
                legend style={
                    at={(0.98, 0.05)}, 
                    anchor=south east, 
                    cells={anchor=west}, 
                    font=\small, 
                    draw=gray!50, 
                    fill=white, 
                    fill opacity=0.8, 
                    text opacity=1,
                    row sep=2pt
                },
            ]
                \addplot[
                    color=gray!80,
                    mark=*,
                    mark options={fill=gray!80, scale=1.2},
                    line width=1.5pt,
                ] coordinates {(10,0.23) (30,0.67) (50,0.71) (80,0.86) (100,0.86)};
                \addlegendentry{BAKU}

                \addplot[
                    color=blue!70!cyan,
                    mark=square*,
                    mark options={fill=blue!70!cyan, scale=1.1},
                    line width=1.5pt,
                ] coordinates {(10,0.45) (30,0.74) (50,0.84) (80,0.87) (100,0.94)};
                \addlegendentry{\name{}}

                
            \end{axis}
        \end{tikzpicture}
        \caption{ }
        \label{fig:libero_curve}
    \end{subfigure}
    \caption{LIBERO benchmark results. (a)~Performance of BAKU (gray) and \name{} (blue) on the LIBERO benchmark. (b)~LIBERO-Long success rate as a function of the fraction of expert demonstrations used for fine-tuning.}
    \label{fig:libero_results}
\end{figure}

\subsection{Results}
\subsubsection{LIBERO Benchmark Results}
\Cref{fig:libero_results} summarizes performance on the LIBERO benchmark.
In \Cref{fig:libero_suites}, we compare \name{} against BAKU~\citep{baku}, a recent state-of-the-art baseline.
Across all four suites, \name{} consistently outperforms BAKU, demonstrating the effectiveness of learning temporally extended latent skills from actionless videos.

\paragraph{LIBERO-Long and data efficiency.}
\Cref{fig:libero_curve} focuses on LIBERO-Long, which consists of long-horizon, multi-stage tasks and therefore provides a more stringent test of whether the learned latent skills capture meaningful temporal structure.
We pretrain a hierarchical policy using pseudo-labels (latent skills and latent actions) extracted by \name{} from diverse actionless videos, and then fine-tune with varying fractions of expert demonstrations.
With only $10\%$ of the demonstrations, BAKU achieves a $23\%$ success rate, whereas \name{} achieves $45\%$, nearly doubling performance.
With $50\%$ of the demonstrations, \name{} reaches $84\%$, comparable to BAKU trained with $100\%$ of the data.
With the full $100\%$ of demonstrations, \name{} achieves $94\%$, outperforming BAKU by a large margin.

\begin{table}[t]
    \centering
    \caption{Ablations on LIBERO-Long, effects of pretraining data and the stage used for latent skill/action conditioning.}
    \label{tab:ablation}
    \small
    \setlength{\tabcolsep}{5.5pt}
    \renewcommand{\arraystretch}{1.05}
    \begin{tabular}{@{}ccccc@{}}
        \toprule
        \shortstack{Method\\~} & \shortstack{Pretraining\\dataset} & \shortstack{Latent\\skill} & \shortstack{Latent\\action} & \shortstack{Success\\rate} \\
        \midrule
        \multirow{4}{*}{BAKU} & Robot & $\text{--}$ & $\bar{z}^{0}$ & $0.87$ \\
                              & Robot & $\text{--}$ & $\bar{z}^{2}$ & $0.81$ \\
                              & Human & $\text{--}$ & $\bar{z}^{0}$ & $0.91$ \\
                              & Human & $\text{--}$ & $\bar{z}^{2}$ & $0.87$ \\
        \addlinespace[2pt]
        \midrule
        \addlinespace[2pt]
        \multirow{7}{*}{\name{}} & $\text{--}$ & $\text{--}$ & $\text{--}$ & $0.67$ \\
                                & Robot & $\bar{z}^{1}$ & $\bar{z}^{0}$ & $0.90$ \\
                                & Robot & $\bar{z}^{2}$ & $\bar{z}^{0}$ & $0.90$ \\
                                & Robot & $\bar{z}^{2}$ & $\bar{z}^{1}$ & $0.87$ \\
                                & Human & $\bar{z}^{1}$ & $\bar{z}^{0}$ & $0.89$ \\
                                & Human & $\bar{z}^{2}$ & $\bar{z}^{0}$ & $\mathbf{0.94}$ \\
                                & Human & $\bar{z}^{2}$ & $\bar{z}^{1}$ & $0.89$ \\
        \bottomrule
    \end{tabular}
\end{table}

\subsubsection{Ablation Studies}
To validate the effectiveness of \name{}, we conduct ablation studies on LIBERO-Long by varying key components, as shown in \Cref{tab:ablation}.
Because \name{} uses a hierarchical policy to jointly leverage latent skills and latent actions, we study which combinations are most effective for pretraining.

First, we vary the pretraining dataset (human vs. robot videos).
Both improve performance, but human videos perform best; we therefore use human pretraining by default.

Next, we vary the stage index $s$ used for conditioning via the unfolded representations $\bar{z}^s$ for latent skills and latent actions.
As shown in \Cref{tab:ablation}, using $s=2$ for latent skills and $s=0$ for latent actions yields the best performance across both human and robot pretraining.
The stage-$2$ representation is produced by the deepest encoder and thus captures longer-range temporal context with more semantically clustered segments; we adopt this setting as default.

We also evaluate whether a non-hierarchical (flat) policy can benefit from latent conditioning by pretraining BAKU with latent actions from either $s=0$ or $s=2$.
While this improves performance, it still lags behind the hierarchical policy, highlighting both the benefit of latent actions and the need for hierarchical policy learning.
Finally, training the hierarchical policy only on the target tasks (without large-scale pretraining) significantly degrades performance, indicating that the gains are not due to the policy architecture alone.

\subsubsection{Dynamic Skill Chunking}
During training, \name{} naturally identifies segment boundaries that correspond to individual, semantically meaningful skills.
As described in \Cref{eq:unfold_map}, we utilize the predicted boundary indicators to partition the trajectory into segments.
Each resulting segment consists of a variable-length sequence of latent actions and is interpreted as a distinct latent skill.

\begin{figure}[t]
    \centering
    \includegraphics[width=\textwidth]{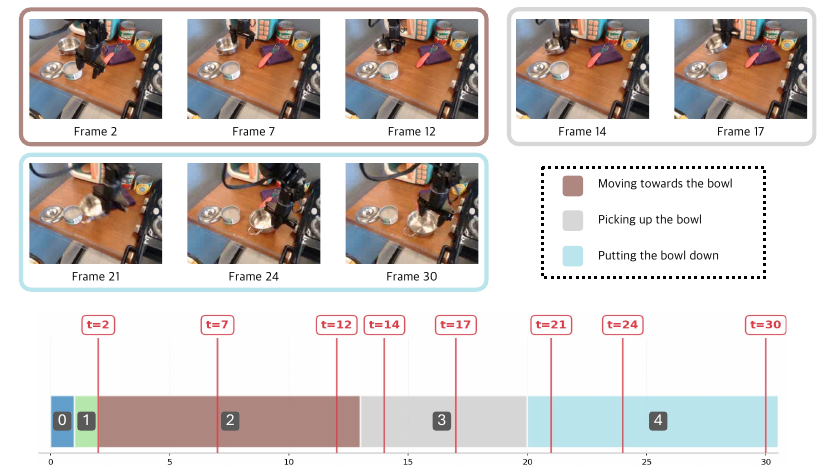}
    \caption{Qualitative results for skill boundary prediction. Using the predicted boundary indicators $b^s_t$, we assign each frame to a skill segment $k^s_t$ and display the segment ID for each segment.}
    \label{fig:fig3}
\end{figure}
To assess this discovery capability, we visualize the inferred chunks in \Cref{fig:fig3}.
Each segment is assigned a distinct color and ID corresponding to the boundary predictions from \name{}.
In Segment 2, the gripper moves across the workspace toward the bowl.
In Segment 3, \name{} predicts a new boundary as the gripper picks up the bowl.
Finally, in Segment 4, the gripper moves to the target location and places the bowl down.

\begin{wrapfigure}{r}{0.5\textwidth}
    \centering
    \vspace{-2em}
    \includegraphics[width=\linewidth]{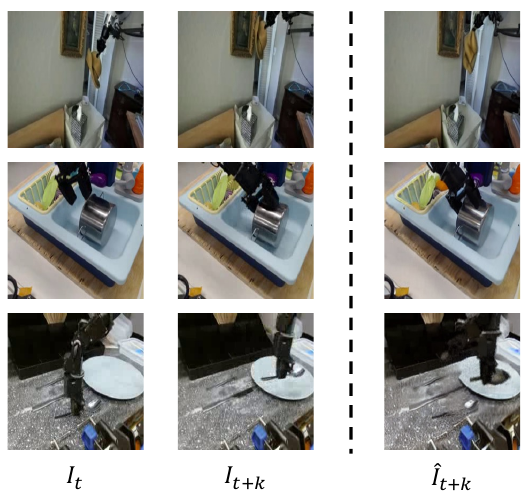}
    \caption{Qualitative results for future frame prediction using a pretrained FDM. Given the current image $I_t$ and the predicted latent action $\hat{z}^l_t$, the model predicts the future frame $\hat{I}_{t+k}$.}
    \label{fig:fig4}
    \vspace{-5em}
\end{wrapfigure}

Despite being trained in a fully unsupervised manner without any labels or ground-truth actions, \name{} consistently groups latent action sequences into coherent skills.
This qualitative result indicates that the proposed dynamic chunking mechanism captures meaningful temporal structure.

\subsubsection{Latent Action Prediction}
As described in \Cref{sec:latent_skill}, \name{} predicts the next latent action from $z^l_t$ to $z^l_{t+1}$.
Therefore, it should preserve the action-like property of the latent representation, i.e., the motion pattern between two frames $I_t$ and $I_{t+k}$.
To verify that the predicted latent actions retain this motion information, we evaluate them via future-frame prediction.
Using the pretrained FDM, we generate the future frame $I_{t+k}$ from the current frame $I_t$ and the predicted latent action $\hat{z}^l_t$.
\Cref{fig:fig4} shows qualitative results.
Although $\hat{z}^l_t$ is predicted from the history $z^l_{:t-1}$, it still yields a consistent future-frame prediction $\hat{I}_{t+k}$.
This indicates that the predicted latent actions retain meaningful motion information, and that \name{} implicitly models temporal dynamics through next-latent prediction.

\section{Conclusion and Limitations}
We present \name{}, a hierarchical latent action model that learns temporally extended latent skills from sequences of low-level latent actions.
Unlike prior work, our approach extracts high-level motion structure from actionless videos without requiring action labels or pre-defined skill sets.
By leveraging a hierarchical architecture with dynamic chunking, \name{} segments variable-length trajectories and encodes each segment into a representative latent skill.
These learned skills improve downstream performance, particularly on long-horizon and multi-stage tasks, while preserving interpretability through next-latent prediction and future frame prediction.
Finally, we demonstrate that using the discovered skills to pretrain a hierarchical policy yields significant data efficiency during fine-tuning.

While our work focuses on encoding latent skills from latent action sequences that capture low-level motion patterns, incorporating language represents a promising direction for future research.
Motion cues and language instructions provide orthogonal rather than parallel information.
Utilizing both signals could lead to a complementary synergy rather than one replacing the other.
For example, imitating motion is often more effective for complex tasks such as furniture assembly, whereas following language instructions can enhance generalizability by imposing fewer constraints compared to specific motion cues.
Therefore, combining hierarchical latent action modeling with natural language is a promising future step.

\paragraph{Limitations.}
While \name{} introduces a novel approach for skill discovery, our experiments are primarily conducted in simulated environments such as LIBERO.
Validating the framework through real-world experiments would further demonstrate the effectiveness of the proposed method.
Additionally, to ensure computational efficiency during temporal modeling, \name{} utilizes a pretrained IDM.
However, training the entire architecture end-to-end could potentially lead to a deeper joint understanding of both low-level motion patterns and high-level skills.

\bibliography{iclr2026_conference}
\bibliographystyle{iclr2026_conference}


\end{document}